\documentclass[11pt]{article}
\usepackage{acl}
\newcommand{\sysname}{Sotto}
\usepackage{microtype}
\usepackage{booktabs}
\usepackage{array}
\usepackage{amssymb}
\usepackage{graphicx}
\usepackage[T1]{fontenc}
\usepackage[utf8]{inputenc}
\usepackage{times}
\usepackage{latexsym}
\graphicspath{{figures/}}

\title{Context-Aware Subtitle Translation on CPU: Deployment-Driven Training, Evaluation, and Latency Design}
\author{Tsz-To Wong \quad Shyang-En Weng \\
  Department of Computer Science \\
  National Yang Ming Chiao Tung University}

\begin{document}
\maketitle
\begin{abstract}
We study English-to-Traditional-Chinese (Taiwan) subtitle translation for a free browser extension, where the model must run on the viewer's CPU in about 500\,ms per subtitle.
We train two models built on the LMT-60 base models, \sysname{}-1.7B and the deployed \sysname{}-0.6B, with a task-specific vocabulary, continued pre-training, supervised fine-tuning, weight interpolation, logit distillation and on-policy preference optimization.
Because reference translations of subtitles are unreliable, we evaluate by pairwise comparison against one fixed baseline, Google Translate, with each pair judged in both orders by a panel of two LLM judges.
On 1,820 held-out subtitles, \sysname{}-1.7B has a win rate of 0.576 against the baseline and wins head-to-head against four open translation models, and is slightly below a 27B model.
We split latency into prefill and decoding: idle-time prefill, which prefills the known part of the prompt while waiting for the next subtitle, cuts mean end-to-end latency by 31\% on a CPU, and the task-specific vocabulary encodes the same Chinese text in 28\% fewer tokens.
The deployed model stays within 500\,ms for every subtitle in our latency test.

\end{abstract}

\section{Introduction}  %
We build English-to-Traditional-Chinese (Taiwan) subtitle translation for a free browser extension that shows a translated line below the platform's official English subtitle.
The extension can read only the subtitle that is on screen, not the subtitle file, so each subtitle must be translated within a fraction of a second after it appears.
There is no backend: the model runs on the viewer's own CPU, with a latency budget of about 500\,ms per subtitle.
Hosted translation is billed per character; at a list price of \$20 per million characters it costs roughly \$0.3--0.65 per hour of viewing, and about three times as much once preceding subtitles are sent as context (Appendix~\ref{app:pricing}).
A product that is free to its users cannot absorb this cost for every active viewer, whereas on-device inference adds no per-request cost for the provider.

\paragraph{Task}
The model translates the current subtitle and also sees the three subtitles before it, the context subtitles, which help to resolve pronouns and ellipsis \citep{voita2018context,tiedemann2017extended}.
The subtitles that follow are not yet available when a subtitle appears.
Section~\ref{sec:system} gives the exact input and output.

\paragraph{What we found}
\emph{Idle-time prefill}, which processes the known part of the next prompt while waiting for the next subtitle, lowers mean end-to-end latency by 31\% on a CPU, and the \emph{task-specific vocabulary} encodes the same Chinese text in 28\% fewer tokens.
With both, \sysname{}-0.6B, a fine-tuned model built on the LMT-60 base models, has a mean latency of 140\,ms on a CPU and a win rate of 0.529 against the baseline, Google Translate, in pairwise comparisons by two LLM judges; one judge puts it slightly below 0.5 and the other above.
\sysname{}-1.7B reaches 0.576, but on the same CPU its latency exceeds 500\,ms for 8.2\% of subtitles, so we offer it as a quality option for viewers with a GPU, where it takes 40\,ms.

\paragraph{Contributions}
(1)~A \emph{latency budget} that splits end-to-end latency into prefill and decoding.
(2)~A \emph{pairwise evaluation protocol} for a task without reliable references: every system is compared with the fixed baseline, and design choices are compared head-to-head.
(3)~A \emph{training recipe} for \sysname{}-0.6B and \sysname{}-1.7B, with an analysis of context, preference optimization and distillation.
(4)~Release of the models, data splits (including the identifiers of the evaluation subtitles) and scripts that rebuild the training data from the original subtitle collection (Ethical Considerations).

\section{Related Work}  %
\paragraph{Subtitle translation}
Subtitling is a distinct translation setting for which customized systems have been built \citep{matusov2019customizing}.
Using preceding text as context for neural translation has been studied, including for movie subtitles \citep{tiedemann2017extended,voita2018context}.
Closest to our setting, \citet{cui2026alpo} align a 14B subtitle translator with preference optimization for batch translation of many subtitles at once.
We translate one subtitle at a time as it appears, with a small model on the viewer's CPU.

\paragraph{Translation with small language models}
Large language models are adapted to translation by continued pre-training followed by a small amount of high-quality parallel fine-tuning \citep{xu2024alma}, and then by contrastive preference optimization (CPO) on translation quality \citep{xu2024cpo}.
Compact multilingual translation models in this style exist at 0.6B and 1.7B parameters \citep{luo2025lmt}, built on Qwen3 \citep{yang2025qwen3}.
We follow this line but add three components suited to a CPU budget: a task-specific vocabulary, knowledge distillation (KD; \citealp{hinton2015distilling}) from the larger model, and weight interpolation after supervised fine-tuning.

\paragraph{Efficiency}
Trimming a multilingual vocabulary to the target language reduces model size while preserving task performance \citep{ushio2023vocabtrim}; we trim and then learn new merges on in-domain text.
Fast translation toolkits such as Marian \citep{junczys2018marian} serve dedicated translation models; we run a decoder-only language model in llama.cpp \citep{gerganov2023llamacpp}.

\paragraph{Evaluation}
Neural metrics trained on human judgments \citep{rei2020comet} and prompted LLMs \citep{kocmi2023gemba} are used to assess translation quality.
LLM judges agree with human preferences on open-ended tasks but show position bias \citep{zheng2023judging,wang2023fair}.
Arena-Hard scores systems by their win rate against a fixed baseline, judges each pair in both orders, and bootstraps confidence intervals \citep{li2024arenahard}.
Our protocol follows this design for a task without reliable references.

\section{System}\label{sec:system}  %
\paragraph{Task}
The input is the current English subtitle and up to three context subtitles that precede it; each is one on-screen subtitle.
The output is the Traditional Chinese (Taiwan) translation of the current subtitle only.
The context subtitles help to resolve pronouns and ellipsis and to fix the tone; they are not translated.
A good output is correct in meaning, natural in Taiwan usage, and concise enough to be read while the subtitle is on screen.

\paragraph{Prompt and vocabulary}
The prompt marks the two fields and the start of the reply with three structural tokens (\texttt{\textbackslash nCTX:\textbackslash n}, \texttt{\textbackslash n\textbackslash nCUR:\textbackslash n} and an assistant token); together they bring the fixed prompt template to about 18 tokens.
Because every output token costs one decode step, we replace the base model's vocabulary with one specific to the task.
We start from the base model's byte-pair-encoding vocabulary \citep{sennrich2016bpe} of 151,669 tokens and remove the tokens for scripts and code that this task does not need and all multi-character Han tokens.
We then learn new merges on Traditional Chinese (Taiwan) subtitle text and remove the least-used remaining base tokens, mostly English fragments, which gives the task-specific vocabulary of 80,000 tokens plus the three structural ones.
Retained tokens keep their embeddings, as do multi-character Han tokens that are learned again; new tokens start from the mean of the embeddings of their constituent base tokens.
On the reference translations of the evaluation set, a subtitle takes 5.43 tokens on average instead of 7.55 under the base vocabulary, 28\% fewer, while English tokenization is almost unchanged: only 7 of 1,820 subtitles are split differently.

\paragraph{Deployment}
The system is a browser extension (Manifest V3) for one streaming platform on macOS and Windows (Table~\ref{tab:deployed}).
A background service worker reads each new subtitle, sends it with the context subtitles to a backend, and shows the translation next to the original.
The native backend is a small helper program on the viewer's device, which the extension reaches through the browser's native messaging API.
It runs llama.cpp \citep{gerganov2023llamacpp} on the CPU, an NVIDIA GPU or Apple Silicon (Metal) at batch size one; llama.cpp supports further GPU backends (Appendix~\ref{app:deploy}).
The model is quantized to Q5\_K\_M, llama.cpp's k-quant format: 5-bit weights with a scale and offset per block of 32, about a quarter at 6 bits (5.8 bits per weight).
For devices that support WebGPU, a second backend runs a 4-bit model in the browser.
The helper caches translations, discards translations of subtitles that a newer subtitle has superseded, and triggers idle-time prefill (Section~\ref{sec:latency}) after each translation.
The helper has no telemetry, and subtitle text stays on the device.
The released extension predates the models in this paper (Appendix~\ref{app:deploy}).

\section{Training Recipe}\label{sec:recipe}  %
Both models follow the same recipe (Figure~\ref{fig:recipe}; settings in Table~\ref{tab:stages} of Appendix~\ref{app:training}), starting from the base checkpoints of the LMT-60 suite \citep{luo2025lmt}, which continue the pre-training of the Qwen3 base models \citep{yang2025qwen3} on monolingual and bilingual text.
Every stage updates all parameters.

\begin{figure*}[t]
\centering
\includegraphics[width=\textwidth]{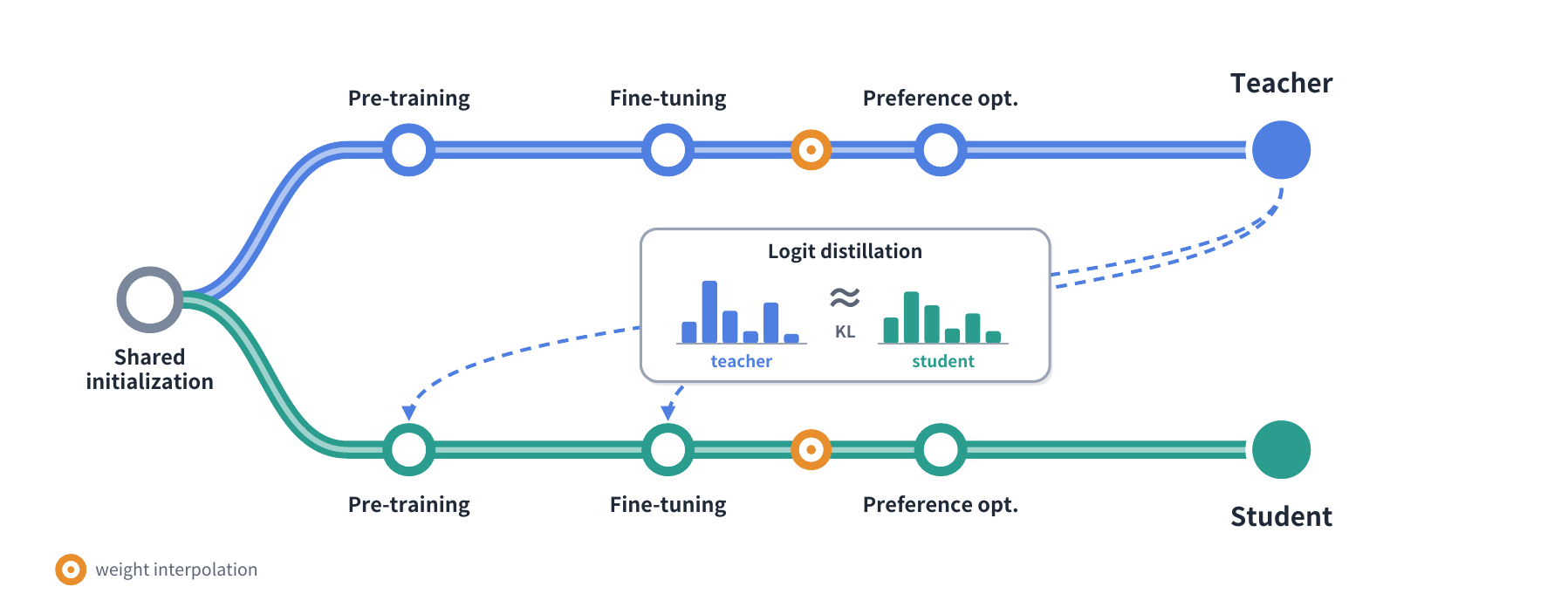}
\caption{Training stages of the two models. \sysname{}-0.6B also learns from \sysname{}-1.7B by logit distillation, in an extra continued pre-training stage and in supervised fine-tuning; weight interpolation follows fine-tuning.}
\label{fig:recipe}
\end{figure*}

\paragraph{Continued pre-training and supervised fine-tuning}
New tokens of the task-specific vocabulary (Section~\ref{sec:system}) start from averaged embeddings, so continued pre-training comes first, on a mixture of Chinese text, English text and bilingual text pairs in both directions.
Supervised fine-tuning uses subtitles selected by an LLM quality filter calibrated on a small human-labeled set (Appendix~\ref{app:data}); the input contains the context subtitles and the current subtitle in the format of Section~\ref{sec:system}, and the loss is computed on the translation only.

\paragraph{Weight interpolation}
Fine-tuning on a narrow dataset can move the weights too far from the starting model.
After fine-tuning, we interpolate the weights as in WiSE-FT \citep{wortsman2022robust}: $\theta=(1-\alpha)\,\theta_{\mathrm{start}}+\alpha\,\theta_{\mathrm{FT}}$, where $\theta_{\mathrm{start}}$ is the continued pre-training model, $\theta_{\mathrm{FT}}$ the fine-tuned one and $\alpha=0.7$, chosen by a sweep that needs no further training.

\paragraph{Logit distillation (\sysname{}-0.6B only)}
In a second continued pre-training stage and in supervised fine-tuning, \sysname{}-0.6B is trained against the full-vocabulary distribution of \sysname{}-1.7B at the same stage, computed online by running the teacher in parallel \citep{hinton2015distilling}.

\paragraph{Preference optimization}
We use contrastive preference optimization (CPO; \citealp{xu2024cpo}), which, unlike direct preference optimization \citep{rafailov2023dpo}, needs no reference model and adds a negative log-likelihood term on the chosen translation.
Preference pairs are built separately for each model; the prompts come from the training split.
The model translates held-out subtitles, and the quality filter finds those it scores poorly.
For each of them we collect candidates (samples from the model, the translation of a larger open model prompted with few-shot examples, and the reference translation) and score them with the filter.
The chosen translation is the best candidate above a threshold that depends on source length, and the rejected one is a clearly worse candidate; the reference translation is never rejected (thresholds in Appendix~\ref{app:data}).
Almost all rejected translations (99\%) are the model's own outputs, which makes the rejected side on-policy, whereas 65\% (\sysname{}-1.7B) and 71\% (\sysname{}-0.6B) of the chosen translations come from the larger open model (Table~\ref{tab:pairs}).
We use the optimized model without a second interpolation.

\section{Evaluation Protocol}\label{sec:protocol}  %
\paragraph{Why pairwise}
Absolute judge scores did not work as our main metric for this task.
An early judge that rewarded completeness disagreed with our own reading of deliberately compressed translations.
Absolute scores also depend on subtitle length.
With one threshold of the LLM quality filter for every length, a preference pair could be built for 44\% of the poorly translated subtitles of 1--3 words but for only 22\% of those with 16 or more words (Appendix~\ref{app:data}).
Negative log-likelihood, our training monitor, can worsen after preference optimization even when quality improves, since the output distribution becomes more peaked.

\paragraph{Pairwise comparison}
A judge sees the context, the current subtitle and two candidate translations, and decides which is more suitable as a subtitle for viewers in Taiwan: meaning first (mistranslations, omissions, wrong register), then naturalness (spoken, concise, wording typical of Taiwan), without favoring length or position.
It returns one of A, B or tie with a one-sentence reason.
Each subtitle is judged twice by each judge, with the order of the two translations swapped.
Identical outputs (ignoring whitespace) are ties and are not sent to a judge.
We use two judges from different vendors, Claude Sonnet 5.5 and Gemini 4 Argon, as a panel \citep{verga2024replacing} (settings in Appendix~\ref{app:judge}) and name no main judge, so that no conclusion rests on one vendor's judgment.
A finding is stated as established only if both judges point the same way.

\paragraph{Scores}
The \emph{preference} for a subtitle is the mean over the two orders and the two judges, counting 1 for A, 0.5 for a tie and 0 for B.
A subtitle is a win, tie or loss for A when its preference is above, equal to or below 0.5.
The \emph{win rate} of A over B is the mean preference over the evaluation set, and 0.5 means no preference.
It equals the mean of the two judges' own win rates; Figure~\ref{fig:scores} illustrates both scores.
We report 95\% bootstrap intervals over subtitles (10,000 resamples; \citealp{koehn2004significance}); resampling groups of near-duplicate documents instead moves the endpoints by at most 0.002 (Table~\ref{tab:cluster}).
As in AlpacaEval \citep{dubois2024lcalpaca}, we define the win rate as a mean of preferences.

\paragraph{Leaderboard and head-to-head}
The \emph{leaderboard} compares every system with the same baseline, Google Translate, so all systems are on one scale and can be ordered.
A \emph{head-to-head} comparison of two named systems addresses one design choice or one competitor and is reported separately from the leaderboard.

\section{Experimental Setup}  %
\paragraph{Data}
Training and evaluation data are English--Traditional Chinese subtitle pairs from the OpenSubtitles collection in OPUS \citep{lison2016opensubtitles}, filtered with rule-based checks (Table~\ref{tab:rules}), a Traditional Chinese script check, and the LLM quality filter; no Simplified-to-Traditional conversion is applied.
The \emph{evaluation set} has 1,820 subtitles from 309 held-out documents that are never used in training: 600 each of 1--3, 4--7 and 8--15 source words and 20 of 16 or more words.
Near-duplicate documents form 247 groups.
Each is exactly one on-screen subtitle according to the time tags of the English side of the corpus and has up to three context subtitles (Appendix~\ref{app:data}).

\paragraph{Systems}
We compare with two commercial services, Google Translate \citep{google2026translate} and DeepL \citep{deepl2026api}, and with the open models Qwen3.8 \citep{qwen2026qwen38}, Breeze 2 \citep{hsu2025breeze2}, Tower+ \citep{rei2025towerplus}, Hunyuan-MT \citep{zheng2025hunyuanmt}, TranslateGemma \citep{translategemma2026} and NLLB-200 \citep{nllb2024}; Table~\ref{tab:leaderboard} gives their sizes.
Each external system uses its vendor's recommended prompt or API settings, and output post-processing is fixed before the comparison (Table~\ref{tab:systems}, Appendix~\ref{app:systems}); no system is given few-shot examples.
Systems that accept context receive the three context subtitles; the others receive only the current subtitle.
The vendor of the second judge also provides the baseline and TranslateGemma; no compared system comes from the vendor of the first.
Two dedicated translation models were excluded because they output Simplified Chinese for 28 and 29 of 30 smoke-test subtitles (Table~\ref{tab:simplified}, Appendix~\ref{app:systems}).

\paragraph{Training data}
Continued pre-training uses about 270M tokens, supervised fine-tuning 208k subtitles kept by the LLM quality filter, and preference optimization 2,991 pairs for \sysname{}-1.7B and 3,749 for \sysname{}-0.6B (Appendix~\ref{app:data}, Table~\ref{tab:pairs}).

\section{Results}\label{sec:results}  %
\begin{table}[t]
\centering\footnotesize\setlength{\tabcolsep}{3pt}
\resizebox{\columnwidth}{!}{%
\begin{tabular}{@{}lclcc@{}}
\toprule
& & \multicolumn{3}{c}{Win rate} \\
\cmidrule(l){3-5}
System & Size & Panel [95\% CI] & Claude & Gemini \\
\midrule
Google Translate & API & 0.500 & -- & -- \\
\midrule
\sysname{}-1.7B & 1.7B & \textbf{0.544} [.527, .560] & .504 & .583 \\
\sysname{}-0.6B & 0.6B & 0.508 [.491, .524] & .478 & .538 \\
Tower+ & 9B & 0.403 [.387, .419] & .414 & .392 \\
Hunyuan-MT & 7B & 0.358 [.342, .374] & .351 & .364 \\
TranslateGemma & 4B & 0.246 [.231, .260] & .239 & .253 \\
NLLB-200 & 3.3B & 0.088 [.078, .098] & .088 & .088 \\
\midrule
DeepL & API & \textbf{0.637} [.622, .652] & .633 & .641 \\
Qwen3.8 & 27B & 0.598 [.582, .614] & .546 & .650 \\
\sysname{}-1.7B & 1.7B & 0.576 [.560, .593] & .536 & .615 \\
\sysname{}-0.6B & 0.6B & 0.529 [.512, .545] & .493 & .565 \\
Breeze 2 & 8B & 0.339 [.322, .356] & .342 & .336 \\
\bottomrule
\end{tabular}}
\caption{Win rate against the baseline, Google Translate, on 1,820 subtitles; \emph{Panel}: mean of the two judges, with 95\% CI. Middle group: current subtitle only; bottom group: with the three context subtitles. Best per group in bold.}
\label{tab:leaderboard}
\end{table}

\paragraph{Leaderboard}
Table~\ref{tab:leaderboard} compares every system with Google Translate.
DeepL, the 27B Qwen3.8 and \sysname{}-1.7B have the highest win rates (0.637, 0.598 and 0.576).
\sysname{}-1.7B is above the baseline under both judges.
\sysname{}-0.6B, the model we deploy, is close to the baseline: 0.529 on the panel, 0.493 and 0.565 under the two judges.
The open models (Tower+, Hunyuan-MT, TranslateGemma, Breeze 2) have lower win rates than the baseline, and NLLB-200 is far below it.

\paragraph{Head-to-head}
Table~\ref{tab:h2h} reports direct comparisons of named systems.
Under both judges, \sysname{}-1.7B has a win rate above 0.5 against \sysname{}-0.6B and against the same model without CPO or without context (Section~\ref{sec:analysis}).
Against external systems, \sysname{}-1.7B beats the four open models under both judges and is slightly below Qwen3.8 at 0.484, with an interval reaching 0.500.
It is below DeepL on the panel (0.456), clearly so under Claude (0.406) and level with it under Gemini (0.506); \sysname{}-0.6B trails both DeepL (0.404) and Qwen3.8 (0.442) under both judges.

\begin{table}[t]
\centering\footnotesize\setlength{\tabcolsep}{3pt}
\resizebox{\columnwidth}{!}{%
\begin{tabular}{@{}lllcc@{}}
\toprule
& & \multicolumn{3}{c}{Win rate} \\
\cmidrule(l){3-5}
A & B & Panel [95\% CI] & Claude & Gemini \\
\midrule
\sysname{}-1.7B & \sysname{}-0.6B & 0.546 [.531, .560] & .537 & .554 \\
\sysname{}-1.7B & w/o CPO & 0.523 [.510, .535] & .525 & .520 \\
\sysname{}-1.7B & w/o context & 0.536 [.525, .548] & .530 & .542 \\
\midrule
\sysname{}-1.7B & DeepL & 0.456 [.441, .472] & .406 & .506 \\
\sysname{}-0.6B & DeepL & 0.404 [.388, .420] & .362 & .447 \\
\sysname{}-1.7B & Qwen3.8 & 0.484 [.469, .500] & .487 & .482 \\
\sysname{}-0.6B & Qwen3.8 & 0.442 [.426, .458] & .456 & .429 \\
\sysname{}-1.7B & Breeze 2 & 0.700 [.684, .716] & .658 & .742 \\
\sysname{}-1.7B & Tower+ & 0.650 [.634, .666] & .609 & .691 \\
\sysname{}-1.7B & Hunyuan-MT & 0.680 [.664, .696] & .644 & .717 \\
\sysname{}-1.7B & NLLB-200 & 0.918 [.908, .928] & .906 & .930 \\
\bottomrule
\end{tabular}}
\caption{Head-to-head win rate of A against B; \emph{Panel}: mean of the two judges, with 95\% CI. A receives the context subtitles; ``w/o context'' removes them at inference; ``w/o CPO'' is the checkpoint before preference optimization.}
\label{tab:h2h}
\end{table}

\begin{figure*}[!t]
\centering
\includegraphics[width=\textwidth]{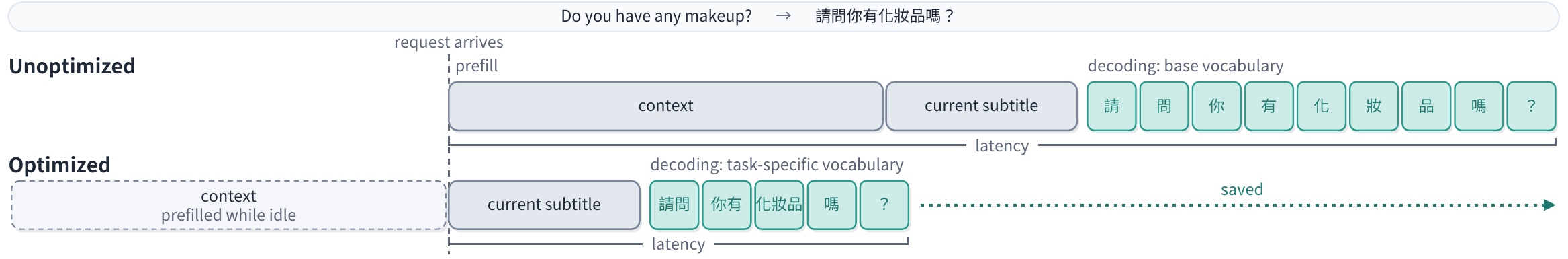}
\caption{Time per subtitle, widths proportional to time. \emph{Unoptimized}: whole prompt prefilled on arrival, base vocabulary, 9 decoding steps. \emph{Optimized}: context prefilled while idle, task-specific vocabulary, 5 decoding steps.}
\label{fig:latency}
\end{figure*}
\section{Analysis}\label{sec:analysis}  %
\paragraph{Context}
With \sysname{}-1.7B, outputs generated with the three context subtitles are preferred over outputs generated without context (win rate 0.536); 1,056 of the 1,820 outputs are identical, 63\% of the comparisons end in a tie, and 62.3\% of the other 673 prefer the output with context.

\paragraph{Preference optimization}
CPO gives \sysname{}-1.7B a win rate of 0.523 against its checkpoint before CPO, with 53.7\% ties, even though held-out negative log-likelihood gets worse (Section~\ref{sec:protocol}).

\paragraph{Logit distillation}
Distilling from \sysname{}-1.7B lowers the held-out loss of \sysname{}-0.6B on bilingual text from 1.424 to 1.373 in continued pre-training, where the teacher reaches 1.318.
In supervised fine-tuning it changes the loss by only 0.008, within the confidence interval of a control run (Appendix~\ref{app:results}).

\paragraph{Subtitle length}
Against the baseline, \sysname{}-0.6B's win rate falls as source length increases (Table~\ref{tab:length}, Appendix~\ref{app:results}): 0.61 for 1--3 words, 0.53 for 4--7 and 0.46 for 8--15.
\sysname{}-1.7B declines less, from 0.63 to 0.51, and the difference between the two models grows with length (\sysname{}-1.7B over \sysname{}-0.6B: 0.52 for 1--3 words, 0.55 for 4--7, 0.57 for 8--15).

\section{Deployment and Latency}\label{sec:latency}  %
The end-to-end latency of a subtitle is the prefill time for the prompt tokens not yet processed plus one decode step per output token; idle-time prefill shortens the first term and the task-specific vocabulary the second (Figure~\ref{fig:latency}).

\paragraph{Protocol}
We run the llama.cpp server with batch size one and the arguments of the shipped helper.
Latency is measured at the server, from receiving a request to its last output token, and excludes the browser and native messaging.
The stream has 897 consecutive subtitles from six real subtitle files with their real arrival times (gaps capped at 4\,s).
We test three devices once each (Appendix~\ref{app:devices}): an Intel Core i5-13500 without a GPU (\emph{CPU}), an Apple M2 with 8\,GB (\emph{Metal}) and an NVIDIA RTX 2080 Ti (\emph{CUDA}).

\paragraph{Ablation}
\emph{Idle-time prefill.}
After a translation, the next prompt is known except for its new subtitle; the client prefills the known part in a zero-output request and keeps the KV cache, so only 12 instead of 42 tokens are prefilled on arrival.
Prefill falls from 152 to 75\,ms (Table~\ref{tab:latency}).
Serving systems reuse the KV cache across requests \citep{kwon2023pagedattention,zheng2024sglang}; our client does so in the gap between subtitles, without changing llama.cpp.
\emph{Task-specific vocabulary.}
The same text needs fewer tokens: decoding falls from 7.1 to 5.3 steps, and each step is cheaper (12.3 instead of 14.7\,ms) because the output layer has 80,003 instead of 151,936 rows.

\begin{table}[t]
\centering\small\setlength{\tabcolsep}{3.5pt}
\begin{tabular}{@{}lrrrrr@{}}
\toprule
& \multicolumn{2}{c}{Prefill} & \multicolumn{2}{c}{Decoding} & \\
\cmidrule(lr){2-3}\cmidrule(lr){4-5}
Configuration & tok. & ms & steps & ms & Mean (p95) \\
\midrule
Base vocab., cold & 42.0 & 152 & 7.10 & 104 & 265 (388) \\
+ idle-time prefill & 12.4 & 75 & 7.12 & 105 & 182 (312) \\
+ task-specific vocab. & 12.4 & 73 & 5.26 & 65 & 140 (228) \\
\bottomrule
\end{tabular}
\caption{Cumulative ablation, \sysname{}-0.6B on the CPU; means in ms per subtitle. \emph{Cold}: the whole prompt is prefilled when the subtitle arrives.}
\label{tab:latency}
\end{table}

\paragraph{Model and device}
\sysname{}-0.6B is the default: 140\,ms on the CPU, with no subtitle above 500\,ms.
\sysname{}-1.7B is a quality option for viewers with a GPU (Table~\ref{tab:devices}).
Decoding reads all weights once per token and is limited by memory bandwidth, so the three times larger model costs 2.3$\times$ as much on the CPU, 1.7$\times$ with Metal and 1.2$\times$ with CUDA.

\begin{table}[t]
\centering\small\setlength{\tabcolsep}{4pt}
\begin{tabular}{@{}lrrr@{}}
\toprule
Model & CPU & Metal & CUDA \\
\midrule
\sysname{}-0.6B & 140 (228) & 96 (138) & 33 (47) \\
\sysname{}-1.7B & 322 (541) & 167 (253) & 40 (61) \\
TranslateGemma-4B$^\ddagger$ & 3,748 (10,712) & -- & -- \\
\bottomrule
\end{tabular}
\caption{Mean (p95) latency in ms with idle-time prefill. On the CPU, 8.2\% of subtitles exceed 500\,ms with \sysname{}-1.7B and all with TranslateGemma; none do elsewhere. $^\ddagger$Current subtitle only, no idle-time prefill; includes waiting for earlier subtitles.}
\label{tab:devices}
\end{table}

\section{Conclusion}  %
We built on-device English-to-Traditional-Chinese (Taiwan) subtitle translation within about 500\,ms per subtitle.
Idle-time prefill cut mean end-to-end latency by 31\%, and a task-specific vocabulary encoded the same text in 28\% fewer tokens.
\sysname{}-1.7B has a win rate above the Google Translate baseline and beats four open translation models; the deployed \sysname{}-0.6B is close to it.

\section*{Limitations}

\paragraph{Motivation rests on assumptions.}
The cost argument uses list prices of commercial machine-translation APIs and character counts per hour measured on our subtitle data (median 23,400 per hour); we have no telemetry and measured no real usage.
Low-cost LLM APIs, which charge per token and may be cheaper than translation APIs priced per character, were not compared; this does not rule out cheaper hosted options.

\paragraph{Early and small deployment.}
The system has a small user base, logs no user data, and has had no user study, so user satisfaction is unmeasured.

\paragraph{Evaluation.}
Quality is measured on one evaluation set of 1,820 subtitles, with LLM judges, for a single language pair and a single target variety, Traditional Chinese (Taiwan).
The set keeps only units that are exactly one on-screen subtitle, so subtitles that the corpus merged from several blocks or split across blocks are absent, and it has almost no subtitles of 16 or more words (20); the reference translations of the 1,409 subtitles added after the first version of the set were not screened by the LLM quality filter.
The human check has two annotators and 120 subtitles in four comparisons (Appendix~\ref{app:human}).
The judges are LLMs and may prefer some styles: they order the systems almost identically, but one gives our models win rates higher by 0.07--0.10, so two comparisons close to parity (\sysname{}-0.6B against the baseline, \sysname{}-1.7B against DeepL) come out differently, and we do not know why (Appendix~\ref{app:judge}).
A commercial system (DeepL) has a higher win rate than \sysname{}-0.6B under both judges, and a large open model is not separated from \sysname{}-1.7B; both are kept in the comparison.
We ran the other systems with their vendors' recommended prompts; two dedicated translation models that output Simplified Chinese were excluded, TranslateGemma, which outputs some Simplified characters (16.5\%), was kept, and post-processing rules were fixed in advance (Appendix~\ref{app:systems}).

\paragraph{Training and measurement.}
Each configuration was trained once, without seed repetition, so we cannot tell whether small differences are real.
End-to-end latency was measured once per configuration on one CPU (i5-13500), one Apple M2 with Metal and one GPU (RTX 2080 Ti, a shared server), with the deployed checkpoints quantized to Q5\_K\_M; other devices will differ, and we report per-token prefill and decoding costs so that others can recompute.

\section*{Ethical Considerations}
The deployed system runs on-device and logs no user data.
Training data are English--Traditional Chinese (Taiwan) subtitle pairs from the OpenSubtitles collection in OPUS \citep{lison2016opensubtitles} (\url{http://www.opensubtitles.org/}, which the collection asks users to link).
The subtitle text belongs to its original rights holders; we claim no rights to it and do not redistribute it.
We will release the model weights, the data splits, document identifiers (including the documents, link identifiers and block rule of the evaluation set), LLM quality filter scores, which translation is chosen in each preference pair, and the scripts that rebuild the training data from the original collection.
No document of the evaluation set is in any training set: the groups of the evaluation set are excluded from the training split, and we checked that no fine-tuning example and no preference prompt comes from them.
The system is a research prototype and is not offered commercially.
Commercial APIs were used as comparison systems and as the judges, and company or product names appear only to identify the systems compared.

\bibliography{refs}

\appendix
\section{Settings of Compared Systems}\label{app:systems}
All systems translated the full evaluation set (1,820 subtitles) at temperature 0; post-processing was fixed before the comparison.
\begin{table}[htbp]\raggedright
\centering\footnotesize\setlength{\tabcolsep}{3pt}
\begin{tabular}{@{}p{0.29\columnwidth}p{0.68\columnwidth}@{}}
\toprule
System & Setting \\
\midrule
Google Translate & Cloud Translation API (basic edition), target zh-TW, current subtitle only (each subtitle translated independently); no post-processing. \\
DeepL & API (free plan), target ZH-HANT, context subtitles passed in the context parameter; no post-processing. \\
Qwen3.8-27B & Task system prompt without examples, context given. \\
Tower+ 9B & Official chat format, ``Translate \ldots{} to Chinese (Traditional)'', current subtitle only; line breaks replaced by spaces. \\
Hunyuan-MT-7B & Official template with Traditional Chinese as target, current subtitle only; line breaks replaced by spaces. \\
TranslateGemma-4B & Official chat template (source en, target zh-TW), current subtitle only, bfloat16, at most 96 new tokens; line breaks replaced by spaces. \\
Breeze 2 8B & Native format with the official system prompt, repetition penalty 1.1, task instruction and context; text after the last \texttt{CUR:} kept. The language model was extracted from the multimodal checkpoint, weights unchanged. \\
NLLB-200 3.3B & Target code zho\_Hant, beam 4, current subtitle only. \\
\bottomrule
\end{tabular}
\caption{Settings of the compared systems.}
\label{tab:systems}
\end{table}
\noindent Two dedicated translation models, GemmaX2-28-9B and Seed-X-PPO-7B, are excluded: their language lists do not include Traditional Chinese, and 29 and 28 of 30 smoke-test outputs, respectively, contained Simplified characters even when Traditional Chinese was requested. We did not convert scripts. On the full evaluation set (GemmaX2 and Seed-X were not run on it), the share of outputs that contain a character a Simplified-to-Traditional converter would change is:
\begin{table}[htbp]
\centering\footnotesize
\begin{tabular}{@{}lc@{}}
\toprule
System & With Simplified characters \\
\midrule
TranslateGemma-4B & 16.5\% \\
Hunyuan-MT-7B & 7.3\% \\
Tower+ 9B & 3.8\% \\
DeepL & 3.6\% \\
Qwen3.8-27B & 3.5\% \\
Breeze 2 8B & 3.5\% \\
\sysname{}-1.7B & 3.1\% \\
Google Translate & 2.9\% \\
\sysname{}-0.6B & 2.4\% \\
NLLB-200 3.3B & 2.0\% \\
\bottomrule
\end{tabular}
\caption{Share of the 1,820 outputs that change under a Simplified-to-Traditional converter (OpenCC \texttt{s2t}); the converter also changes some variant forms that occur in Traditional text, so the lowest values are not zero.}
\label{tab:simplified}
\end{table}
\noindent TranslateGemma lists Traditional Chinese for Taiwan as a target, but 16.5\% of its outputs contain such characters, more than twice the share of any other system; we did not convert them, so its win rate may be understated.

\section{Judges}\label{app:judge}
\begin{figure*}[t]
\centering
\includegraphics[width=0.85\textwidth]{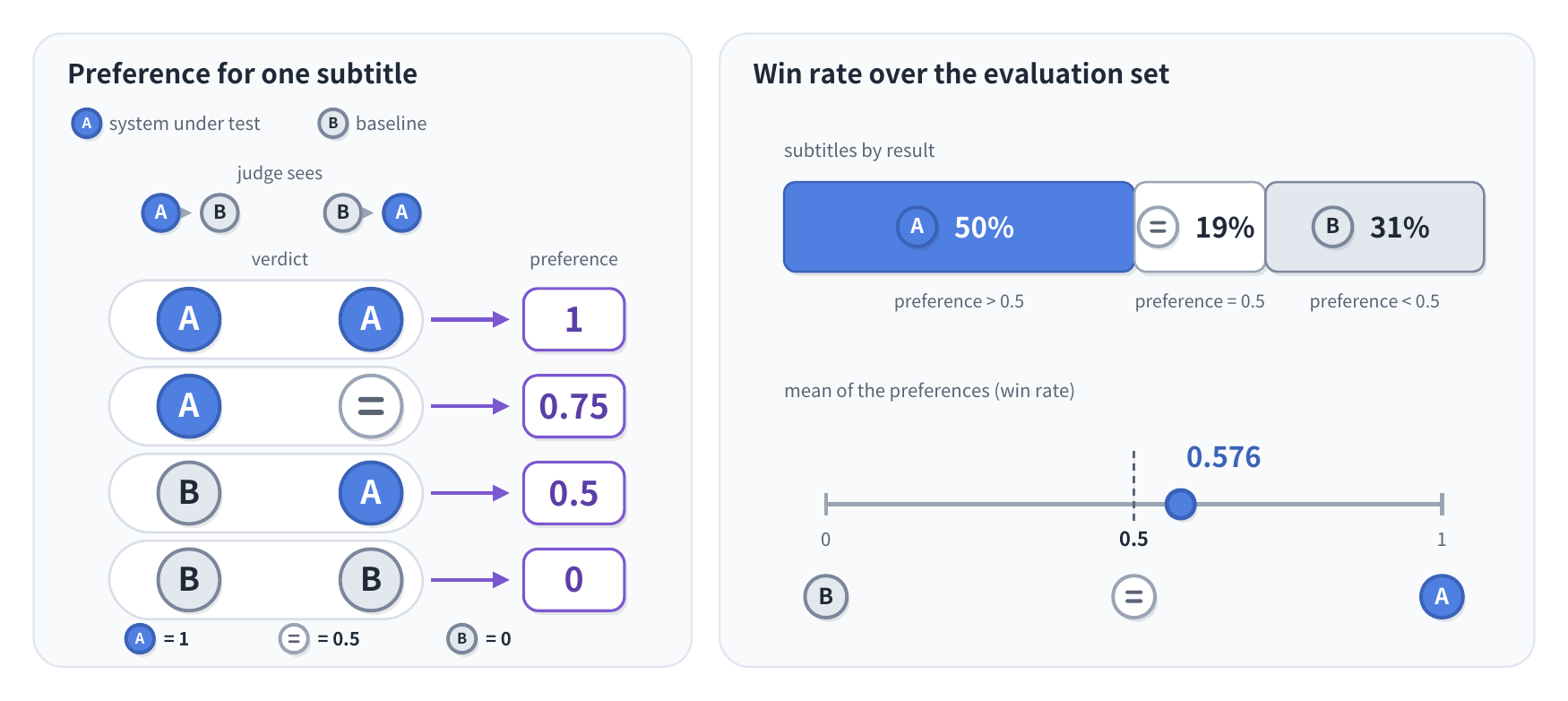}
\caption{How the scores are read. Left: a judge sees each subtitle twice with the order of the two translations swapped; its two verdicts give a preference of the system under test (1, 0.75, 0.5, 0.25 or 0), and with two judges the four verdicts are averaged. Right: over the evaluation set, subtitles are wins, ties or losses by their preference, and the win rate is the mean preference (here \sysname{}-1.7B against the baseline).}
\label{fig:scores}
\end{figure*}
All pairwise comparisons were judged by two LLM judges in October 2026, each with the same system prompt and user message (Figure~\ref{fig:judgeprompt}), which returns a one-sentence reason and a verdict as JSON.
The judge receives the context subtitles, the current subtitle and the two translations; each pair is judged in both orders, and pairs with identical outputs are ties without a call.
\emph{Claude Sonnet 5.5} (\texttt{claude-sonnet-5-5}) was run through the vendor's batch interface.
The model accepts no temperature or other sampling setting, so sampling uses the defaults and the outputs are not deterministic; extended thinking is switched off with the lowest setting the model offers, the output is limited to 400 tokens, and the system prompt is cached between requests.
\emph{Gemini 4 Argon} (\texttt{gemini-4-argon}) was run through the vendor's batch prediction service with default sampling and the lowest thinking level.
Each judge ran each comparison once.

\begin{figure*}[t]
\newcommand{\qq}{\textquotedbl}
\fbox{\begin{minipage}{\dimexpr\textwidth-2\fboxsep-2\fboxrule\relax}\footnotesize
\textbf{System prompt}\par\smallskip
You are a senior subtitle reviewer in Taiwan. You are given the preceding English subtitles \texttt{CTX} (at most 3), the English subtitle to be translated \texttt{CUR}, and two candidate Chinese translations \texttt{A} and \texttt{B}. Decide which one is more suitable to be shown as it is, as a Traditional Chinese subtitle, to a Taiwanese audience.

\smallskip Order of judgment:
\begin{enumerate}\setlength{\itemsep}{0pt}\setlength{\parsep}{0pt}\setlength{\topsep}{2pt}
\item \textbf{Meaning}: any mistranslation, any omission of key information, a wrong negation or question tone, wrong terminology or forms of address, an idiom translated literally; anything added that is not in the source. Use \texttt{CTX} to resolve references and ellipsis. The side with a clear error of meaning loses, unless the other side's error is more serious.
\item \textbf{Naturalness}: does it read like a subtitle written by a Taiwanese translator? Colloquial, concise wording with the right tone (including profanity and sentence-final particles) is a plus; English sentence patterns, translationese, bureaucratic style, mainland Chinese usage and word-for-word rendering are minuses.
\item If both are correct in meaning and both are natural, compare tone and concision; choose \texttt{tie} if the difference is small.
\end{enumerate}
Notes:
\begin{itemize}\setlength{\itemsep}{0pt}\setlength{\parsep}{0pt}\setlength{\topsep}{2pt}
\item Do not prefer a translation because it is longer or shorter. Compressing or paraphrasing a subtitle is a plus, as long as no important information is lost.
\item Do not prefer a translation because of its position, A or B.
\item If both have errors, choose the one with the lighter error; choose \texttt{tie} only if they are identical or hard to tell apart.
\item Leftover untranslated English words count against a translation (except names and proper nouns that are usually kept in the original).
\end{itemize}
Output (a single line of JSON, fields in this fixed order): \texttt{reason} is one sentence (at most 20 words) naming the deciding difference; \texttt{winner} is \texttt{\qq A\qq}, \texttt{\qq B\qq} or \texttt{\qq tie\qq}.\\
\texttt{\{\qq reason\qq: \qq A drops the negation of wouldn't\qq, \qq winner\qq: \qq B\qq\}}

\medskip\textbf{User message}\par\smallskip
\texttt{\#\#\# CTX}\\
\texttt{[1] }\textit{context subtitle 1}\quad\texttt{[2] }\ldots\quad\texttt{[3] }\ldots\quad or \texttt{(none)} without context\\
\texttt{\#\#\# CUR}\\
\textit{current subtitle}\\
\texttt{\#\#\# A}\\
\textit{first translation}\\
\texttt{\#\#\# B}\\
\textit{second translation}\\
\texttt{Output the JSON.}
\end{minipage}}
\caption{Judge prompt, used unchanged for both judges. In the user message each context subtitle is on its own line; the second call of each pair swaps A and B.}
\label{fig:judgeprompt}
\end{figure*}

\paragraph{How the judges compare.}
Table~\ref{tab:judges} lists the win rate under each judge and on the panel.
Over the 22 comparisons the win rates of the two judges correlate at 0.97 (mean absolute difference 0.044, largest 0.104), and the judges order 54 of the 55 pairs of systems in the leaderboard the same way; the one pair they swap is DeepL and Qwen3.8 (0.633 and 0.546 with Claude, 0.641 and 0.650 with Gemini).
In 18 of the 22 comparisons they point the same way (or both find no difference); the other four are \sysname{} against Google Translate or DeepL, with win rates within 0.1 of 0.5 under both judges.
Gemini gives \sysname{} and Qwen3.8 win rates 0.07--0.10 higher than Claude does; the judges agree on the comparisons with context, preference optimization and model size.
We do not know why they differ in level.
\begin{table}[htbp]
\centering\footnotesize\setlength{\tabcolsep}{4pt}
\resizebox{\columnwidth}{!}{%
\begin{tabular}{@{}lcccc@{}}
\toprule
 & Claude & Gemini & Panel & Win/tie/loss (\%) \\
\midrule
DeepL & 0.633 & 0.641 & 0.637 & 51.8/26.4/21.8 \\
Qwen3.8 & 0.546 & 0.650 & 0.598 & 53.6/16.7/29.7 \\
\sysname{}-1.7B & 0.536 & 0.615 & 0.576 & 50.1/18.7/31.2 \\
\sysname{}-0.6B & 0.493 & 0.565 & 0.529 & 44.9/19.5/35.5 \\
\sysname{}-1.7B, no context & 0.504 & 0.583 & 0.544 & 47.7/18.2/34.1 \\
\sysname{}-0.6B, no context & 0.478 & 0.538 & 0.508 & 43.5/18.4/38.1 \\
Tower+ & 0.414 & 0.392 & 0.403 & 25.4/27.3/47.3 \\
Hunyuan-MT & 0.351 & 0.364 & 0.358 & 23.7/21.0/55.2 \\
Breeze 2 & 0.342 & 0.336 & 0.339 & 24.2/16.9/58.9 \\
TranslateGemma & 0.239 & 0.253 & 0.246 & 15.0/15.9/69.1 \\
NLLB-200 & 0.088 & 0.088 & 0.088 & 5.5/3.1/91.4 \\
\midrule
\sysname{}-1.7B vs. DeepL & 0.406 & 0.506 & 0.456 & 36.0/22.1/41.9 \\
\sysname{}-0.6B vs. DeepL & 0.362 & 0.447 & 0.404 & 31.3/19.9/48.8 \\
\sysname{}-1.7B vs. Qwen3.8 & 0.487 & 0.482 & 0.484 & 36.5/25.6/37.9 \\
\sysname{}-0.6B vs. Qwen3.8 & 0.456 & 0.429 & 0.442 & 32.6/24.1/43.3 \\
\sysname{}-1.7B vs. \sysname{}-0.6B & 0.537 & 0.554 & 0.546 & 34.3/40.7/25.0 \\
\sysname{}-1.7B vs. w/o CPO & 0.525 & 0.520 & 0.523 & 26.1/53.7/20.2 \\
\sysname{}-1.7B vs. w/o context & 0.530 & 0.542 & 0.536 & 23.0/63.0/14.0 \\
\sysname{}-1.7B vs. Breeze 2 & 0.658 & 0.742 & 0.700 & 65.5/13.2/21.3 \\
\sysname{}-1.7B vs. Tower+ & 0.609 & 0.691 & 0.650 & 59.3/17.5/23.2 \\
\sysname{}-1.7B vs. Hunyuan-MT & 0.644 & 0.717 & 0.680 & 62.7/16.7/20.5 \\
\sysname{}-1.7B vs. NLLB-200 & 0.906 & 0.930 & 0.918 & 91.9/3.1/4.9 \\
\bottomrule
\end{tabular}}
\caption{Win rate of the first-named system under each judge and on the panel, with the panel's win/tie/loss shares. Top: against Google Translate; Tower+, Hunyuan-MT, TranslateGemma and NLLB-200 receive no context. Bottom: head-to-head.}
\label{tab:judges}
\end{table}

\section{Data Pipeline}\label{app:data}
\begin{figure*}[t]
\centering
\includegraphics[width=0.9\textwidth]{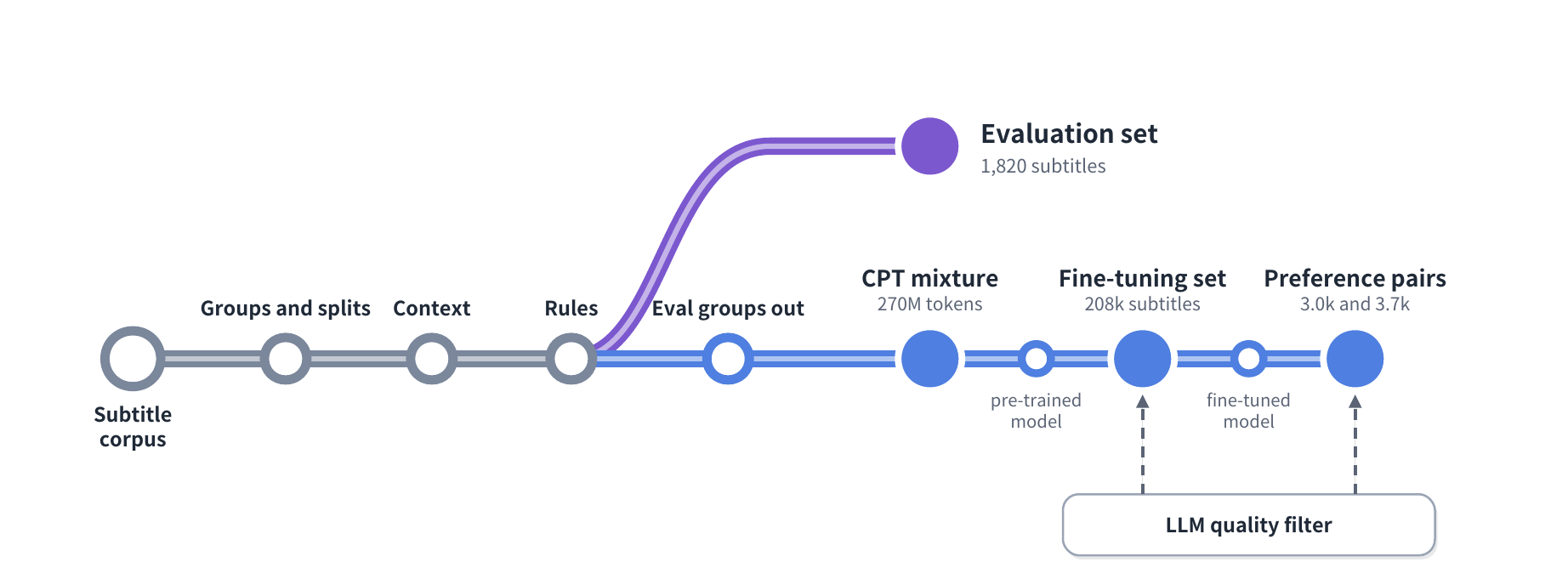}
\caption{Data pipeline. The evaluation set branches off first and its groups are removed from everything after it. The training sets follow in the order of training: preference pairs are built from the outputs of the fine-tuned model, and the LLM quality filter selects the fine-tuning data and finds the poor outputs. The numbered paragraphs below follow the steps.}
\label{fig:datapipe}
\end{figure*}

This appendix (Figure~\ref{fig:datapipe}) lists every step from the raw corpus to the three training sets (continued pre-training, supervised fine-tuning, preference optimization) and the evaluation set.
Counts are from the runs behind the paper; steps that are deterministic use fixed seeds.

\paragraph{1. Corpus and ingestion.}
We use a public collection of aligned movie and TV subtitles (release of 2024) with English--Traditional Chinese alignment.
It has two parts: a development part of 1,314 documents (808,084 aligned subtitles) whose plain-text files carry no timestamps (documents are separated by blank lines whose positions agree in the two languages), and a main part of 22,689 documents (18.6M aligned subtitles) with timestamps.
For the main part, we download the alignment links, the Chinese side of every aligned document, and, from the English archive, only the aligned English documents (about 22.7k files, fetched individually); plain-text versions are matched to the links in order.
A public table of series episodes adds series ids.
All source files are recorded with content hashes in a snapshot manifest.

\paragraph{2. Groups and splits.}
Near-duplicate documents (re-releases, episodes of one series, copied subtitles) must fall into the same split.
We shingle the English text of each document into word 5-grams, ignore shingles that occur in more than 2\% of documents (advertisements, credits), find candidate pairs with MinHash (128 permutations, 32 bands of 4 rows) and keep an edge when the exact Jaccard similarity is at least 0.40; episodes of one series are merged through the series id.
Connected components are groups.
Held-out splits are drawn only from the development part, by group, as shares of its subtitles: 90\% train, 2\% validation, 4\% development mining (inspected repeatedly) and 4\% frozen test (59 documents, 36,746 subtitles; never used for training, filtering, tuning or the results of this paper).
All main-part documents are training data, except those that share a group with a held-out development document, which are used for neither training nor evaluation.
The seed is fixed (20260929) and the split is stored as an immutable manifest.

\paragraph{3. Examples with context.}
For every aligned subtitle, the context subtitles are the up to three preceding subtitles of the same document.
In the main part the chain breaks at a silence longer than 5,000\,ms or when a start time goes backward by more than 500\,ms; the development part has no timestamps, so only the document start and the window limit the context.
In the training data of the main part 72.5\% of the examples have three context subtitles, 7.4\% two, 8.9\% one and 11.3\% none (27.2\% of the chains end at a gap); in the development part 99.5\% have three.

\paragraph{4. Cleaning, rules and target style.}
Cleaning removes invisible direction and zero-width characters, leaked subtitle-format tags and empty brackets, and the outer square brackets of lyric subtitles when the English has none; the raw text is kept in its own column.
Rule flags are computed on the cleaned text and nothing is deleted; a row is \emph{kept} when it passes all of the following:
\begin{itemize}\itemsep0pt
\item not empty;
\item no character that occurs only in Simplified Chinese, and no leftover Simplified-style variants;
\item no garbled text (mojibake);
\item no subtitle-credit or advertisement pattern (Chinese or English);
\item no common English words left in the Chinese subtitle, no square brackets that the English lacks, no merged two-speaker dash subtitles;
\item not untranslated: a subtitle with at least 12 Latin letters or 3 Latin words must have a share of Chinese characters of at least 0.34;
\item length limits: at most 80 English words and 160 Chinese characters, Chinese-to-English character ratio between 0.05 and 3.0, and no long Chinese target for an English subtitle shorter than 10 characters;
\item document-level: a document is dropped as a whole when at least 10\% of its subtitles are Simplified;
\item no more than five occurrences of an identical (English, Chinese) pair within a split;
\item no overlap: a training subtitle whose English (at least 6 words) also occurs in validation, development-mining or frozen-test data is dropped.
\end{itemize}
\begin{table}[htbp]
\centering\footnotesize
\begin{tabular}{@{}lrr@{}}
\toprule
Rows failing & Development & Main \\
\midrule
Simplified-only character & 4.8\% & 14.9\% \\
Garbled text & 0.3\% & 3.0\% \\
English left in Chinese & 0.7\% & 1.2\% \\
Untranslated & 0.7\% & 1.5\% \\
Simplified document rule & 6.0\% & 22.6\% \\
Repeated pairs & 5.3\% & 8.3\% \\
Kept after all rules & 86.5\% & 67.9\% \\
Rows (training split) & 721,731 & 18,574,547 \\
\bottomrule
\end{tabular}
\caption{Effect of the rules on the training split (the shares of different rules overlap).}
\label{tab:rules}
\end{table}
Finally, the Chinese side is normalized in style, for every subtitle: half-width sentence punctuation becomes full-width, the sentence-final mark of the English (question mark, exclamation mark, ellipsis) is copied when missing, breaks that the translator already placed between clauses are written as a single space, no full stop is added at the end, and ellipses are written as \texttt{……}; no new clause break is ever inserted.

\paragraph{5. Difficulty tags.}
Rule-based tags mark phenomena such as idioms, pragmatic or figurative language, sarcasm, profanity, insults, emotional subtitles, colloquial or phrasal verbs, negation, entities, numbers, multi-clause subtitles, high semantic density, long subtitles, and subtitles likely to need omission or context.
About 74--79\% of rows carry at least one broad tag, so a \emph{strict} subset (idiom, pragmatic, figurative, sarcasm, profanity, insult, high semantic density, long subtitle, omission-prone; about 14\% of the kept rows of the main part) is used to oversample hard cases.

\paragraph{6. Earlier fine-tuning candidates.}
Candidates come only from the training split, kept rows, and rows that are not part of an audit sample.
Two pools per part: \emph{natural}, uniform over all eligible rows, and \emph{targeted}, rows with a strict tag; a cap per document (400 for the development part, 40 for the main part) keeps long films from dominating.
The first pool has 344,937 candidates (development natural 141,000 and targeted 58,937; main natural 94,000 and targeted 51,000).
An earlier LLM judge with a completeness-oriented rubric and four yes/no flags kept those without any flag (249,498 rows); a second pool of 600,167 candidates was judged only in part before we stopped (185,936 judged), and its kept rows (136,978) were added.
This earlier fine-tuning set has 386,476 rows (source length 1--3, 4--7, 8--15 and 16 or more words: 21.2, 36.1, 33.6 and 9.1\%), with all groups of the evaluation set excluded.
We later found that this judge favored literal, complete translations over compressed ones, which is why we re-scored the rows with the quality filter below.

\paragraph{7. Evaluation set.}
The evaluation set is drawn from the held-out documents of the main part (the development part of the earlier version, which has neither time tags nor alignment ids, is not used).
A candidate is an aligned unit whose English side is exactly one on-screen subtitle, that is, its English sentences contain one start and one end time tag; units that join several blocks (110,292 of 328,394) or are fragments of a block are dropped, as are units without a Chinese counterpart, units whose reference translation has Simplified-only characters (165), units without letters or with a music note, and units with an empty side.
Of the 2,000 subtitles of the earlier version (screened by two LLM annotators with a written rubric, which flagged 129 whose reference translation is unsuitable), 411 satisfy the rule and are kept unchanged, with their context; the new rows are sampled uniformly within a bucket from the held-out documents of that version to 600 subtitles for each of the source-length buckets 1--3, 4--7 and 8--15 words, with at most 12 per group (seed 20261020).
The new rows were not screened by the quality filter.
Subtitles of 16 or more words are 0.6\% of the units and are kept only from the earlier version (20 subtitles).
The current subtitle is unique in the set after normalization; the context is the previous three aligned units of the same file, with the chain broken at a silence longer than 5\,s or a start time that goes backward by more than 500\,ms (the rule reproduces the context of all 728 main-part rows of the earlier version).
Every group that contributes a subtitle is held out: all its pool rows are removed from every training set.
The result has 1,820 subtitles from 309 documents and 247 groups: 1,722 natural and 98 targeted rows (all reused), with three context subtitles for 1,276 of them (two, one and none for 159, 172 and 213).

\paragraph{8. Long pool, quality filter and fine-tuning set.}
The earlier set has few long subtitles (the filter below keeps only about 30\% of the rows with 8--15 words and 20\% of those with 16 or more), so we add a pool of 186,603 subtitles with 8--22 words from the training split that are in no earlier pool, fine-tuning set, preference-prompt pool, evaluation set or evaluation document, with all three context subtitles, at most 25 rows per document, and structural pre-filters (between 0.9 and 3.0 Chinese characters per English word, at most one dialogue dash and two sentence boundaries, no garbled text, no leftover Simplified-looking characters, no Latin words in the target); the pool is shuffled so that partial runs are unbiased.
We score the 573,079 candidates (the earlier set and the long pool) with an LLM quality filter (Qwen3.8-27B \citep{qwen2026qwen38} in an INT4 quantization, served with vLLM, temperature 0) that sees the English subtitle, its context subtitles, the neighboring subtitle pair from the same file, and the reference translation, and decides whether the translation is a good training example: correct in meaning, fluent, and natural as Taiwan subtitles.
It outputs a grade, 2 (keep), 1 (marginal) or 0 (reject), and two flags that say why (a meaning problem, a style problem).
Compression, omission and free translation are not penalized; mistranslation, wrong tone, translationese and mainland usage are.
The \emph{keep probability} of a row is the probability that the filter's grade is 2: the probability of the token ``2'' at the grade position, normalized over the tokens ``0'', ``1'' and ``2'' (from the top five log-probabilities), so it is a continuous score of how likely the filter is to keep the row.
We calibrated the rubric on 60 subtitles that one annotator (an author) labeled as keep (34), marginal (9) or reject (17), which also serve as the few-shot examples in the filter's prompt.
In two-fold cross-validation (the filter sees only the labels of the other half), keep-versus-not accuracy was 77\%, and a threshold of 0.8--0.9 on the keep probability kept about 87--89\% of rows that the annotator also kept, at a recall of about half; with only 60 labels these figures are rough.
We chose purity over volume: 214,209 rows (37.4\%) have a keep probability of at least 0.9 (42.9\% of the earlier set, 26.0\% of the long pool).
We then drop rows that contain mojibake, invisible direction marks, leftover Simplified-looking characters or Latin words in the target, and exact duplicates of (source, target), which leaves 207,716 rows: 159,363 from the earlier set and 48,353 from the long pool.
By source words, 43,945 rows have 1--3, 65,986 have 4--7, 88,300 have 8--15 and 9,485 have 16 or more; 187,729 have three context subtitles, 5,646 two, 6,439 one and 7,902 none.

\paragraph{9. Continued pre-training mixture.}
The mixture has three sources with the same token budget (about 100M tokens each): monolingual Chinese, monolingual English and bilingual pairs.
Monolingual English is every kept English subtitle of the training split; monolingual Chinese is the unique kept Chinese subtitles of the training split with a per-document cap, without Simplified-only characters and without evaluation groups.
Subtitles are joined with newlines and cut into sequences of 256 tokens, with the loss on every token (288,125 Chinese and 393,852 English sequences).
The bilingual pool has 3M pairs from the same filtered split with a cap of 600 pairs per document, without context subtitles (the context format is learned in fine-tuning); rows that the earlier judge kept (Step 6) are always included and the rest is a seeded random sample.
A pair is written in one of two directions with equal probability, as \texttt{<en2zh\_tw> English} then \texttt{<zh\_tw> Chinese} or the reverse with \texttt{<zh\_tw2en>} and \texttt{<en>}, followed by the end marker; several pairs are packed into one sequence of 256 tokens without cross-sample attention masking, and the loss covers the target side only (356,191 sequences; 28.9\% of their tokens carry loss, 9.2\% are padding).
Altogether the mixture has about 1.04M sequences, or 270M tokens, and both models use it.

\paragraph{10. Preference pairs.}
Pairs are built separately for each model from a pool of 150,000 subtitles with three context subtitles from the training split (disjoint from the evaluation set: none of its 18,227 documents belongs to a group of the evaluation set), of which 30,000 are screened:
(1)~the model translates every screening prompt greedily and the quality filter scores the output;
(2)~outputs scored below 0.3 become the prompts to improve, at most 8,000, after dropping garbled sources and music-note subtitles (849, 3,019, 3,403 and 729 prompts for sources of 1--3, 4--7, 8--15 and 16 or more words);
(3)~for each such prompt we collect candidates: the greedy output, four samples from the model (temperature 0.8, top-$p$ 0.95), one translation by a larger open model prompted with eight few-shot examples and the formatting rules of the target text (no commas or full stops, sentence-final marks following the English source), and the reference translation;
(4)~the filter scores every distinct candidate;
(5)~the chosen translation is the best-scoring candidate if it passes the source-length threshold (0.90, 0.85, 0.75 and 0.60 for 1--3, 4--7, 8--15 and 16 or more words), the rejected one the worst-scoring candidate (never the reference translation) if it scores below 0.3, the score gap is at least 0.5, and ties are broken at random;
(6)~a pair is dropped if either side has garbled text, the chosen translation has Simplified-only characters or loses a number of the source, or the length ratio of chosen to rejected is outside 0.6--1.5.
With one threshold of 0.9 for every length, pairs could be built for 44\%, 40\%, 34\% and 22\% of the prompts of the four length buckets (in a run with an earlier policy model); with the length-dependent thresholds, for 45\%, 48\%, 53\% and 59\%.
\begin{table}[htbp]
\centering\footnotesize
\begin{tabular}{@{}lrr@{}}
\toprule
 & \sysname{}-1.7B & \sysname{}-0.6B \\
\midrule
Pairs & 2,991 & 3,749 \\
\midrule
Chosen from the larger model & 1,954 & 2,655 \\
Chosen from samples & 630 & 503 \\
Chosen from the reference & 407 & 591 \\
\midrule
Rejected from samples & 2,055 & 3,018 \\
Rejected from the greedy output & 917 & 718 \\
Rejected from the larger model & 19 & 13 \\
\midrule
Source of 1--3 words & 235 & 304 \\
Source of 4--7 words & 1,042 & 1,263 \\
Source of 8--15 words & 1,370 & 1,751 \\
Source of 16 or more words & 344 & 431 \\
\midrule
Chosen longer than rejected & 42\% & 40\% \\
\bottomrule
\end{tabular}
\caption{Preference pairs of the two models, including the 300 pairs of each that are held out (training uses 2,691 and 3,449).}
\label{tab:pairs}
\end{table}

\section{Training Pipeline}\label{app:training}
\paragraph{Vocabulary.}
Starting from the 151,669-token vocabulary of the base model (151,643 merged tokens and 26 special tokens), we remove 35,996 tokens for scripts and code that this task does not use (never seen in the training text or in the protected held-out text), delete the 16,407 tokens that contain more than one Han character together with their dependents, add 30,000 merges learned on Traditional Chinese (Taiwan) subtitle text, and remove the 49,266 remaining base tokens least used in the English and Chinese training text (protected tokens excluded), which leaves 80,000; then we add three structural tokens, which gives 80,003: \texttt{\textbackslash nCTX:\textbackslash n}, \texttt{\textbackslash n\textbackslash nCUR:\textbackslash n} and \texttt{assistant}.
The second removal is mostly English fragments (37,923, e.g.\ \texttt{.php}, \texttt{https}) and Simplified Chinese single characters; every token that remains keeps its string and merge, and all 256 byte tokens are kept, so any text can still be encoded.
A kept token keeps its embedding; of the 30,000 merges, 4,080 recreate a deleted multi-character Han token, which gets its original embedding back, and the other 25,920 are new tokens that start from the mean of the embeddings of their parts; the output layer is tied to the embeddings.
On the reference translations of the evaluation set this changes the average from 7.55 to 5.43 tokens per subtitle (1.28 to 1.78 characters per token), while English tokenization is almost unchanged (7 of the 1,820 current subtitles are split differently, token ratio 1.0002); the embedding matrix (151,936 rows in the base model) has about 74M fewer parameters in the 0.6B model and 147M fewer in the 1.7B model, and the fixed prompt template takes about 18 tokens.

\paragraph{Stages and settings.}
Table~\ref{tab:stages} lists the training stages in order; all runs use two 24\,GB GPUs (RTX 4090) with ZeRO and gradient checkpointing.
Between the stages: the embeddings of the base model are remapped to the 80,000-token vocabulary, and a warm-up trains only the rows of the new tokens; after continued pre-training three structural tokens are added, each row initialized as the mean of the rows of the fragments it replaces; after fine-tuning the weights are interpolated with $\alpha=0.7$ toward the model after continued pre-training (for \sysname{}-1.7B a sweep over 0.5--0.9 had its best held-out loss at 0.6).
Preference optimization uses CPO with $\beta=0.1$, NLL weight 1.0, warm-up 10\%, AdamW, bf16, maximum length 256 and 300 pairs held out, and its result is used as is.
\begin{table*}[t]
\centering\footnotesize\setlength{\tabcolsep}{3.5pt}
\begin{tabular}{@{}llllrrr@{}}
\toprule
Stage & Model & Data & Loss & Steps & Learning rate & Batch \\
\midrule
Warm-up (new token rows only) & both & pre-training mixture & CE & 300 & $5\times10^{-4}$ & 192 \\
Continued pre-training & 1.7B & pre-training mixture, 1 epoch & CE & 5,407 & $1.5\times10^{-5}$ & 192 \\
Continued pre-training & 0.6B & pre-training mixture, 1 epoch & CE & 5,407 & $2\times10^{-5}$ & 192 \\
Logit distillation, pre-training & 0.6B & pre-training mixture, 0.75 epoch & $0.9\cdot$KL$+0.1\cdot$CE & 6,082 & $3\times10^{-5}$ & 128 \\
Supervised fine-tuning & 1.7B & fine-tuning set, 2 epochs & CE & $\approx$1,080 & $2.8\times10^{-5}$ & 384 \\
Logit distillation, fine-tuning & 0.6B & fine-tuning set, 2 epochs & $0.7\cdot$KL$+0.3\cdot$CE & $\approx$1,080 & $2.8\times10^{-5}$ & 384 \\
Preference optimization & 1.7B & 2,691 pairs, 3 epochs & CPO & 126 & $3\times10^{-6}$ & 64 \\
Preference optimization & 0.6B & 3,449 pairs, 3 epochs & CPO & 159 & $3\times10^{-6}$ & 64 \\
\bottomrule
\end{tabular}
\caption{Training stages in order. Batch is the global batch (sequences or examples); CE is cross-entropy; the teacher of each distillation stage is described below.}
\label{tab:stages}
\end{table*}

\paragraph{Logit distillation.}
The student's loss is computed against the teacher's full-vocabulary distribution, with the frozen teacher running next to the student on the same batch.
The teacher of each stage is the 1.7B model of the same stage, not the final one: after continued pre-training for the pre-training stage, after fine-tuning and interpolation (before preference optimization) for fine-tuning.
We use the full vocabulary online because storing only the teacher's top-$k$ probabilities covers only about 60\% of the probability mass for Chinese ($k=64$), so the long tail is not transferred.

\section{Human Check}\label{app:human}
This is a preliminary consistency check of the panel: on 120 subtitles, the annotators' win rates are close to the panel's.
Two native speakers of Taiwan Mandarin, who were not told the source of either translation, compared translations of 120 subtitles in four comparisons (240 judgments), choosing A, B or tie by the criteria of Section~\ref{sec:protocol}.
The subtitles are those whose two outputs differ by at least two characters after removing punctuation, 10 from each length bucket, with the positions of A and B random.
Table~\ref{tab:human} gives the win rate of the first-named system on these subtitles, as the mean of the two annotators and under each judge.
Over the four comparisons the annotators' win rate is 0.050 above the panel's (95\% interval $-0.015$ to $0.114$), so the two cannot be told apart on this sample; it is 0.108 above Claude's alone and 0.008 below Gemini's alone.
The two annotators agree with each other on 48\% of the subtitles, and each judge agrees with each annotator on 44--52\%.
\begin{table*}[t]
\centering\footnotesize\setlength{\tabcolsep}{3pt}
\begin{tabular}{@{}lcccc@{}}
\toprule
Comparison (first named wins) & Humans & Claude & Gemini & Panel \\
\midrule
\sysname{}-1.7B vs. Google Translate & 0.583 & 0.583 & 0.658 & 0.621 \\
\sysname{}-0.6B vs. Google Translate & 0.533 & 0.433 & 0.517 & 0.475 \\
\sysname{}-1.7B vs. DeepL & 0.492 & 0.333 & 0.550 & 0.442 \\
\sysname{}-0.6B vs. DeepL & 0.475 & 0.300 & 0.392 & 0.346 \\
\midrule
All four (120 subtitles) & 0.521 & 0.412 & 0.529 & 0.471 \\
Humans minus column [95\% interval] & --- & $+0.108$ [0.042, 0.177] & $-0.008$ [$-0.083$, 0.065] & $+0.050$ [$-0.015$, 0.114] \\
\bottomrule
\end{tabular}
\caption{Human check: win rate of the first-named system on 30 subtitles per comparison, 120 in total. \emph{Humans} is the mean of the two annotators.}
\label{tab:human}
\end{table*}

\section{Additional Results}\label{app:results}
Held-out loss of \sysname{}-0.6B on bilingual text falls from 1.424 to 1.373 with logit distillation during continued pre-training (the \sysname{}-1.7B teacher: 1.318). In supervised fine-tuning, the KD loss changes the held-out loss by 0.008, within the confidence interval of a control.
\begin{table}[htbp]\raggedright
\centering\footnotesize
\begin{tabular}{@{}lccc@{}}
\toprule
 & 1--3 & 4--7 & 8--15 \\
\midrule
\sysname{}-1.7B & 0.633 & 0.587 & 0.510 \\
\sysname{}-0.6B & 0.605 & 0.532 & 0.455 \\
\bottomrule
\end{tabular}
\caption{Win rate of the panel against Google Translate by source length (600 subtitles per bucket; the 20 subtitles of 16 or more words are not shown).}
\label{tab:length}
\end{table}

Win rates with group-level intervals, and on the natural rows only (the other 98 rows are targeted rows reused from the earlier evaluation set):
\begin{table}[htbp]
\centering\footnotesize\setlength{\tabcolsep}{3pt}
\resizebox{\columnwidth}{!}{%
\begin{tabular}{@{}llcc@{}}
\toprule
System & Context & All rows (1,820) & Natural rows (1,722) \\
\midrule
DeepL & \checkmark & 0.637 [0.622, 0.652] & 0.639 [0.624, 0.654] \\
Qwen3.8 & \checkmark & 0.598 [0.582, 0.614] & 0.599 [0.583, 0.615] \\
\sysname{}-1.7B (ours) & \checkmark & 0.576 [0.560, 0.591] & 0.578 [0.562, 0.594] \\
\sysname{}-0.6B (ours) & \checkmark & 0.529 [0.511, 0.546] & 0.535 [0.517, 0.552] \\
Breeze 2 & \checkmark & 0.339 [0.321, 0.356] & 0.342 [0.324, 0.360] \\
\sysname{}-1.7B (ours) & $\times$ & 0.544 [0.528, 0.560] & 0.549 [0.533, 0.566] \\
\sysname{}-0.6B (ours) & $\times$ & 0.508 [0.490, 0.525] & 0.511 [0.492, 0.529] \\
Tower+ & $\times$ & 0.403 [0.385, 0.421] & 0.406 [0.387, 0.424] \\
Hunyuan-MT & $\times$ & 0.358 [0.340, 0.375] & 0.359 [0.341, 0.376] \\
TranslateGemma & $\times$ & 0.246 [0.231, 0.261] & 0.251 [0.236, 0.266] \\
NLLB-200 & $\times$ & 0.088 [0.078, 0.099] & 0.091 [0.080, 0.102] \\
\bottomrule
\end{tabular}}
\caption{Win rate of the panel against Google Translate with 95\% intervals that resample the 247 groups of near-duplicate documents (10,000 resamples), for all rows and for the natural rows only.}
\label{tab:cluster}
\end{table}

\section{Cost Arithmetic}\label{app:pricing}
The list price of Google Cloud Translation is \$20 per million characters (\url{https://cloud.google.com/translate/pricing}; checked October 2026).
A subtitle of the evaluation set has 31.2 source characters on average, so about \$0.0006 per subtitle. The 3,010 documents of the corpus with reliable timestamps (at least 99\% of the cues tagged, at least 800 aligned units each) have a median of 23,400 source characters per hour of viewing (10th--90th percentile 14,800--32,400; mean 23,600), so about \$0.47 per hour (\$0.30--0.65); the same documents have a median of 759 aligned units per hour.
Sending the three context subtitles adds about three times as many characters, and about three times as much cost if context is billed.

\section{Deployment Details}\label{app:deploy}
\begin{table*}[t]
\centering\footnotesize
\begin{tabular}{@{}p{0.16\textwidth}p{0.80\textwidth}@{}}
\toprule
Item & Setting \\
\midrule
Platform & Chrome-based browsers; macOS and Windows 10/11 (x64) \\
Native backend & Helper program on the device, started on demand through native messaging; llama.cpp on CPU, NVIDIA CUDA or Apple Silicon (Metal); model quantized to Q5\_K\_M \\
Browser backend & 4-bit (\texttt{q4f16\_1}) model, about 272\,MB, 64,024-token vocabulary, 256-token context window, 16 output tokens; needs WebGPU with 16-bit shader support, no CPU fallback \\
Request & Temperature 0, up to 24 output tokens, stop at the end-of-turn token; 10\,s timeout \\
Idle-time prefill request & Sent after a translation: the next prompt without the generation suffix, 0 output tokens, prompt cache on; requests are serialized and only the latest pending one is sent; errors are ignored \\
Service layer & In-memory cache keyed by model, prompt version, current subtitle and context; identical in-flight requests share one call; a result is discarded if a newer subtitle has arrived \\
Data & No telemetry, analytics or crash reports; subtitle text is not sent to a hosted service; the network is used to install and to download models \\
\bottomrule
\end{tabular}
\caption{Deployed system.}
\label{tab:deployed}
\end{table*}
\paragraph{Released build and the models of this paper.}
The build available at submission uses an earlier model (model version 2026.07) whose 64,024-token vocabulary was trained as a new byte-pair-encoding vocabulary and has since been replaced by the task-specific vocabulary, and one context subtitle by default (0--8 allowed); the models evaluated in this paper use the 80,003-token vocabulary and three context subtitles.
\paragraph{Backend comparison.}
On the project's built-in workload of 100 subtitles and the earlier model (Q5\_K\_M for the native helper), on one development machine (Windows, 12 logical processors, NVIDIA RTX 3080), the mean (p95) end-to-end latency was 47 (85)\,ms for the browser backend, 147 (287)\,ms for the native helper on CPU and 28 (49)\,ms with CUDA.
These numbers use a different workload and model from Section~\ref{sec:latency} and are not comparable with it.
\paragraph{Hardware backends.}
llama.cpp \citep{gerganov2023llamacpp} lists 17 backends in its documentation (checked October 2026), with their target devices: BLAS (all), BLIS (all), CANN (Ascend NPU), CUDA (Nvidia GPU), HIP (AMD GPU), Hexagon (Snapdragon), IBM zDNN (IBM Z and LinuxONE), MUSA (Moore Threads GPU), Metal (Apple Silicon), OpenCL (Adreno GPU), OpenVINO (Intel CPUs, GPUs and NPUs; marked as in progress), RPC (all), SYCL (Intel GPU), VirtGPU (VirtGPU APIR), Vulkan (GPU), WebGPU (all) and ZenDNN (AMD CPU), in addition to the CPU path.
We ran the native helper with the Metal backend on an Apple M2 and measured the latency of the models of this paper there, as on an Intel Core i5-13500 CPU and an NVIDIA RTX 2080 Ti with CUDA (Section~\ref{sec:latency}).
WebGPU is an API that browsers implement on top of native graphics APIs; the Dawn and wgpu implementations list Direct3D 12, Metal, Vulkan and OpenGL as their native backends (their documentation, checked October 2026).
Our browser backend uses WebGPU and has no CPU fallback.

\section{Latency Devices}\label{app:devices}
We measured one device for each backend of the native helper (Table~\ref{tab:hw}).
All runs use the server arguments of the shipped helper (a 512-token context window and the default number of threads of llama.cpp; with Metal and CUDA all layers run on the GPU), one run per model and device.
\paragraph{Why these devices.}
Decoding reads all weights once per output token, so memory bandwidth more than compute sets the cost of a step (Section~\ref{sec:latency}); the three devices span about an order of magnitude of it, from dual-channel desktop memory (tens of GB/s) through the 100\,GB/s of the M2 to the 616\,GB/s of the 2080 Ti.
Each sits at the lower end of its backend rather than at the top.
The CPU is a mid-range desktop processor with no GPU, the path that every viewer can use.
The M2 with 8\,GB is the entry configuration of Apple Silicon, where the model shares memory with the operating system, the browser and the video.
The 2080 Ti is a consumer card from 2018, not a data-center accelerator.
A model that meets the latency budget on these devices therefore does not depend on high-end hardware.
No machine was idle or tuned for the measurement. The CPU machine ran Linux with other services and the power-saving frequency governor; the M2 ran macOS on mains power with other processes, at a load average of up to 23 on 8 cores; the 2080 Ti sits in a shared Linux server where another user's job ran on a second GPU.
This is closer to a viewer's computer, where the browser and the video player run next to the model, than a dedicated benchmark host is.
The devices do not cover laptop processors on battery power, Windows, or AMD and Intel GPUs, and other devices will differ (Limitations).
\begin{table}[htbp]
\centering\footnotesize\setlength{\tabcolsep}{3pt}
\begin{tabular}{@{}>{\raggedright\arraybackslash}p{0.14\columnwidth}>{\raggedright\arraybackslash}p{0.26\columnwidth}>{\raggedright\arraybackslash}p{0.31\columnwidth}>{\raggedright\arraybackslash}p{0.20\columnwidth}@{}}
\toprule
Backend & Device & Compute & Memory \\
\midrule
CPU & Intel Core i5-13500 desktop, 2023 & 14 cores (6 performance, 8 efficiency), 20 threads, up to 4.8\,GHz; no GPU used & 16\,GB system memory \\
Metal & Apple M2, 2022 & 8 CPU cores (4 performance, 4 efficiency) and the integrated GPU & 8\,GB unified, 100\,GB/s \\
CUDA & NVIDIA GeForce RTX 2080 Ti, 2018, in a server with an Intel Xeon W-2125 & 4,352 CUDA cores; host CPU with 4 cores, 8 threads & 11\,GB GDDR6, 616\,GB/s \\
\bottomrule
\end{tabular}
\caption{Devices of the latency measurements (Section~\ref{sec:latency}). The same quantized model files, with identical SHA-256, run on all three; llama.cpp builds b11548 (CPU), b11789 (Metal) and a local build (CUDA).}
\label{tab:hw}
\end{table}

\end{document}